\documentclass{article}

\usepackage{amsmath,amsthm,amssymb,type1cm}
\usepackage[dvipdfm]{graphicx,color}
\usepackage{bm}
\usepackage{url}
\usepackage{float}	

\title{A Bayesian estimation approach to analyze non-Gaussian data-generating processes \\ with latent classes}

\author{Naoki Tanaka\thanks{The Institute of Scientific and Industrial Research, Osaka University} \and Shohei Shimizu$^*$ \and Takashi Washio$^*$}

\date{}

\begin{document}
\maketitle

\begin{abstract}
A large amount of observational data has been accumulated in various fields in recent times, and there is a growing need to estimate the generating processes of these data. A linear non-Gaussian acyclic model (LiNGAM) based on the non-Gaussianity of external influences has been proposed to estimate the data-generating processes of variables. However, the results of the estimation can be biased if there are latent classes. In this paper, we first review LiNGAM, its extended model, as well as the estimation procedure for LiNGAM in a Bayesian framework. We then propose a new Bayesian estimation procedure that solves the problem. 
\end{abstract}

\section{Introduction}
Several methods have recently been proposed to discover a complete causal structure, that is, all the causal directions, under the assumption that disturbance variables have non-Gaussian distributions. However, the estimation results can be biased if there are "latent classes." Latent classes are unobserved discrete variables that have more than one observed child variables. Data that has been generated from different processes are mixed in the presence of latent classes. Several methods have been proposed to estimate the causal structure in the presence of latent classes \cite{Shimizu07ICONIP}, but all of these are affected by local optima. Therefore, in this paper, we propose a new estimation approach that can solve this problem.

The structure of this paper is as follows. In Section \ref{既存手法}, we briefly review the data generating process for estimating causal structure, (LiNGAM, short for Linear Non-Gaussian Acyclic Model) \cite{Shimizu06JMLR}, the LiNGAM mixture model \cite{Shimizu07ICONIP}, existing estimation approaches to the LiNGAM mixture model \cite{Shimizu07ICONIP}, and the BayesLiNGAM model. In Section \ref{提案手法}, we extend previous research dedicated to solving the problem, and test the performance of our new method through experiments with simulated data in Section \ref{評価実験}. Section \ref{結論} concludes the paper.

\section{Background}
\label{既存手法}

	\subsection{Linear non-Gaussian acyclic model (LiNGAM)}
	\label{LiNGAMモデル}
\begin{figure}[h]
			\centering
			\includegraphics[width=80mm,clip]{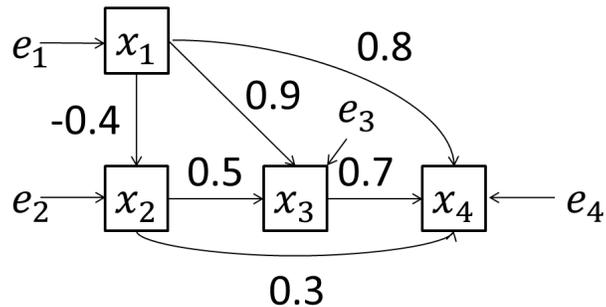}
			\caption{An example of LiNGAM described by a DAG.}
			\label{LiNGAMのDAG}
		\end{figure}

		We begin by introducing the basic LiNGAM model \cite{Shimizu06JMLR}. LiNGAM is a causal model with the following four assumptions.
		\begin{enumerate}
			\item The relations between the observed variables $x_i (i = 1, \dots, n)$ can be represented by a directed acyclic graph (DAG), as shown in Fig. \ref{LiNGAMのDAG}.
			\item $x_i$ is assigned a value by a {\sl linear function} of the values already assigned to the variables constituting its parents in the DAG as well as a "disturbance" (noise) term $e_i$ and an optimal constant term $\mu_i$:
				  \begin{equation}
					\label{LiNGAMモデルの式}
					x_i = \sum_{k(j) < k(i)} b_{ij} (x_j - \mu_j) + e_i + \mu_i
				  \end{equation}
			where $k(i)$ is a causal order of $x_i$ in the DAG (so that if there is a directed edge from $x_j$ to $x_i$ in the DAG, $k(j) < k(i)$), $b_{ij}$ represents the strength of the connection between $x_j$ and $x_i$ in the DAG.
			\item The $e_i$ are all continuous random variables that follow {\sl non-Gaussian} distributions with {\sl zero means} and non-zero variances, and the $e_i$ are mutually independent, i.e., $p(e_1, \dots, e_n) = \prod_{i} p_i(e_i)$.
			\item The dataset $\mathbf{D} = \{\mathbf{x}^1, \dots, \mathbf{x}^N\}$ (each $\mathbf{x}$ contains components $x_i$) are observed, and each data vector $\mathbf{x}$ is generated according to the process described above with the same DAG, coefficients $b_{ij}$, constants $\mu_i$, and disturbances $e_i$, sampled independently from the same distributions.
		Note that this assumption implies that there are either no unobserved (latent) confounders \cite{Pearl00book} (hidden variables) or that the $x_i$ are causally sufficient \cite{Spirtes93book}.
		\end{enumerate}

	\subsection{LiNGAM mixture model}
	\label{LiNGAM混合モデル}
		The LiNGAM mixture model \cite{Shimizu07ICONIP} is an extension of the basic LiNGAM to cases where the final assumption of the basic LiNGAM model collapses because of the presence of latent classes (hidden groups). If they exist, each latent class has a different structure. "Structure" here consists of the causal order $k(i)$, the connection strengths $b_{ij}$, the densities of disturbances $p_i$, and the means of observed variables $\mu_i$. Therefore, structure in a class $c$ can be described using $k^{(c)}(i)$, $b_{ij}^{(c)}$, $p_i^{(c)}$, and $\mu_i^{(c)}$ (Note that the disturbances $e_i^{(c)}$ are generated from $p_i^{(c)}$). The data within each class ($c = 1, \dots, l$) are assumed to be generated by the basic LiNGAM. Thus, the data-generating model within class $c$ can be described as follows:
		\begin{equation}
			\label{LiNGAM混合モデルの式}
			x_i = \sum_{k^{(c)}(j) < k^{(c)}(i)}^{} b_{ij}^{(c)} (x_j - \mu_j^{(c)}) + \mu_i^{(c)} 
			+ e_i^{(c)}
		\end{equation}
		Moreover, the data vectors $\mathbf{x}$ are assumed to be generated by the following mixture density:
		\begin{equation}
			\label{確率密度の混合}
			p(\mathbf{x} | \mathbf{\Theta}) = \sum_{c = 1}^{l} \biggl\{ p\bigl( \mathbf{x} | \mbox{\boldmath $\theta$}^{(c)} \bigr)\biggr\} P(c) \\
		\end{equation}
		where $l$ is the number of classes, and $\mathbf{\Theta} = [\mbox{\boldmath $\theta$}^{(1)}, \dots, \mbox{\boldmath $\theta$}^{(l)}]$, and $\mbox{\boldmath $\theta$}^{(c)}$ contain all the parameters in (\ref{LiNGAM混合モデルの式}), i.e., $k^{(c)}(j), k^{(c)}(i), b_{ij}^{(c)}, \mu_j^{(c)}, \mu_i^{(c)}$, and $p_i^{(c)}(e_i^{(c)})$. Note that if only one latent class exists, the LiNGAM mixture model is equivalent to basic LiNGAM.

	\subsection{Existing estimation approach to LiNGAM mixture model}
	\label{Amica}
		In this subsection, we briefly introduce the existing estimation approach to the LiNGAM mixture model. In this approach, the LiNGAM mixture model is transformed into an Independent Component Analysis (ICA) \cite{Hyva01book}) model, and an ICA algorithm (for example, \cite{Mollah-mixtures} \cite{Palmer08ICASSP}) is applied to it \cite{Shimizu07ICONIP}.
		
		The LiNGAM mixture model (Equation (\ref{LiNGAM混合モデルの式})) can be written in matrix form as follows:
		\begin{equation}
			\label{LiNGAM混合モデルの式（行列）}
			\mathbf{x} = \mathbf{B}^{(c)} \mathbf{x} + (\mathbf{I} - \mathbf{B}^{(c)}) 
			\mbox{\boldmath $\mu$}^{(c)} + \mathbf{e}^{(c)}
		\end{equation}
		where $\mathbf{B}^{(c)}$, $\mbox{\boldmath $\mu$}^{(c)}$, and $\mathbf{e}^{(c)}$ collect $b_{ij}^{(c)}$, $\mu_i^{(c)}$, and $e_i^{(c)}$, respectively, and $\mathbf{I}$ is the identity matrix. We can then obtain the following equation by solving Equation (\ref{LiNGAM混合モデルの式（行列）}) for $\mathbf{x}$:
		\begin{equation}
			\label{ICA混合モデルの式（行列）}
			\mathbf{x} = \mbox{\boldmath $\mu$}^{(c)} + \mathbf{A}^{(c)} \mathbf{e}^{(c)}
		\end{equation}
		where $\mathbf{A}^{(c)} = (\mathbf{I} - \mathbf{B}^{(c)})^{-1}$.
		We can estimate $\mathbf{B}^{(c)}$ by calculating $\mathbf{A}^{(c)}$ using the ICA algorithm and computing $\mathbf{B}^{(c)}$.
		However, there are two important indeterminacies that ICA cannot solve: the order and the scaling of the independent components.
		In order to solve them, we need to permute and normalize $\mathbf{A}^{(c)}$ appropriately before using it to compute $\mathbf{B}^{(c)}$ (See \cite{Shimizu06JMLR} for more details).
	

	\subsection{The BayesLiNGAM method}
	\label{BayesLiNGAM}
		Existing estimation methods for the LiNGAM mixture model \cite{Mollah-mixtures} \cite{Palmer08ICASSP} are affected by local optima, because of which we propose a Bayesian approach. In this subsection, we summarize the BayesLiNGAM method \cite{Hoyer09UAI}, which is an estimation method of basic LiNGAM.

		In Bayesian inference, the inference with the highest posterior probability for some hypotheses is selected. In BayesLiNGAM, the hypotheses are different possible DAGs. BayesLiNGAM outputs the DAG with the highest posterior probability of all possible DAGs given the data. The posterior probabilities can be calculated by Bayes' theorem:
		\begin{equation}
			\label{ベイズの定理}
			P(G_m|\mathbf{D}) = \frac{p(\mathbf{D}|G_m)P(G_m)}{p(\mathbf{D})}
		\end{equation}
		Here, $G_m$ are the different possible DAGs, $m = \{1, \dots, N_g\}$, where $N_g$ denotes the number of different DAGs on $n$ variables, and $\mathbf{D}$ is the observed dataset （$\mathbf{D} = \left\{\mathbf{x}^1, \dots ,\mathbf{x}^N\right\}$, $N$ : sample size). Due to our assumption of independent and identically distributed (IID) data, $p(\mathbf{D}) = \prod_{s = 1}^{N} p(\mathbf{x}^s)$. We then need to specify the likelihoods $p(\mathbf{D}|G_m)$, prior probabilities $P(G_m)$, and normalization constant $p(\mathbf{D})$ to compute the posterior probabilities.

The prior probabilities $P(G_m)$ first incorporate any domain of knowledge and prior information. If we have no knowledge or information about the $G_m$, all values of $P(G_m)$ are equal, i.e., $P(G_m) = \frac{1}{N_g}$.

		Furthermore, $p(\mathbf{D})$ is a constant that simply normalizes posterior probabilities. Hence, we can obtain $p(\mathbf{D})$ as follows:
		\begin{equation}
			\label{事後確率の正規化定数}
			P(\mathbf{D}) = \sum_{m = 1}^{N_g} p(\mathbf{D}|G_m) P(G_m)
		\end{equation}

		The likelihoods $p(\mathbf{D}|G_m)$ can be computed by marginalizing over $p(\mathbf{D}|\mathbf{\Theta},G_m)$ of $\mathbf{\Theta}$:
		\begin{equation}
			\label{周辺尤度}
			P(\mathbf{D}|G_m) = \int p(\mathbf{D}|\mathbf{\Theta}, G_m)p(\mathbf{\Theta}|G_m)d\mathbf{\Theta}
		\end{equation}
 		Here, $\mathbf{\Theta}$ collects all the parameters in Equation (\ref{LiNGAMモデルの式}) (i.e., $k(j), k(i), b_{ij}, \mu_j, \mu_i$, and $p_i$), $p(\mathbf{D}|\mathbf{\Theta}, G_m)$ denotes the likelihood of the model (here, basic LiNGAM), and $p(\mathbf{\Theta}|G_m)$ denotes the prior distributions of $\mathbf{\Theta}$. Note that if the $G_m$ are given, we can find the parents of $x_i$ in the $G_m$ and determine the causal order $k(j)$ and $k(i)$. That is, the first term of the right side in Equation (\ref{LiNGAMモデルの式}) can be specified by the $G_m$.
		We introduce the likelihood of the model and prior distribution for $\mathbf{\Theta}$ in Section \ref{提案手法}.

\section{Proposed method}
\label{提案手法}
	\subsection{Model}
	\label{データ生成モデル}
		We assume that the data within each class ($c = 1, \dots, l$) are generated by the basic LiNGAM. Thus, the data-generating model in class $c$ can be described by Equation (\ref{LiNGAM混合モデルの式}), and the probability densities of the data $\mathbf{x}$ are the same as in Equation (\ref{確率密度の混合}). We then use the Bayesian approach to estimate the LiNGAM mixture model in the same manner as BayesLiNGAM in Section \ref{BayesLiNGAM}.

		Here, we need to parameterize the densities $p_i$ to specify $p(\mathbf{D}|\mathbf{\Theta},G_m)$ in Equation (\ref{周辺尤度}). Due to the assumption in LiNGAM that the $e_i$ are all continuous random variables with {\sl non-Gaussian} distributions, we use a generalized Gaussian distribution \cite{Nadarajah05} that incorporates a shape parameter. The generalized Gaussian distribution is symmetric and includes Gaussian, Laplacian, continuous uniform, and several non-Gaussian distributions.
		The probability density function is as follows:
		\begin{equation}
			\label{一般化ガウス分布}
			p_i(e_i) = \frac{\lambda_i\sqrt{\frac{\Gamma(3/\lambda_i)}{\Gamma(1/\lambda_i)}} {\rm exp} (-(\sqrt{\frac{\Gamma(3/\lambda_i)}{\Gamma(1/\lambda_i)}}\frac{|e_i|}{\sigma_i})^{\lambda_i})}{2\sigma_i\Gamma(1/\lambda_i)}
		\end{equation}
		Here, $\sigma_i$ represent the standard deviations, $\lambda_i$ are the shape parameters, and $\Gamma()$ denotes the Gamma function.

		Moreover, we need to transform the density of the $e_i$ into that of the $x_i$. The LiNGAM mixture model (Equation (\ref{LiNGAM混合モデルの式})) can be written in matrix form as follows:
		\begin{eqnarray}
			\label{x,eの関係式（行列）}
			\nonumber \mathbf{x} &=& \mathbf{f}(\mathbf{e}^{(c)}) \\
                                                  &=& \mbox{\boldmath $\mu$}^{(c)} + (\mathbf{I} - \mathbf{B}^{(c)})^{-1} \mathbf{e}^{(c)}
		\end{eqnarray}
		where $\mathbf{f}()$ is a mapping vector, and $\mathbf{B}^{(c)}$, $\mathbf{e}^{(c)}$, and $\mbox{\boldmath $\mu$}^{(c)}$ collect $b_{ij}^{(c)}$, $e_i^{(c)}$, and $\mu_i^{(c)}$, respectively. $\mathbf{I}$ is the identity matrix.
		
		The density of $\mathbf{x}$ within class $c$ $\bigl(p(\mathbf{x} | \mbox{\boldmath $\theta$}^{(c)},G_m)\bigr)$ is obtained from the density of $\mathbf{e}^{(c)}$ $\bigl(p_\mathbf{e}^{(c)}(\mathbf{e}^{(c)})\bigr)$ as follows \cite{Hyva01book}:
		\begin{eqnarray}
			\label{確率密度の変数変換(1)}
			p(\mathbf{x} | \mbox{\boldmath $\theta$}^{(c)},G_m) = \frac{1}{|{\rm det} J\mathbf{f}(\mathbf{f}^{-1}(\mathbf{x}))|} p_\mathbf{e}^{(c)}(\mathbf{f}^{-1}(\mathbf{x}))
		\end{eqnarray}
		where $J\mathbf{f}$ is a Jacobian matrix.
		From equations (\ref{x,eの関係式（行列）}) and (\ref{確率密度の変数変換(1)}) with the assumption of acyclicity, ${\rm det} J\mathbf{f}(\mathbf{f}^{-1}(\mathbf{x}))$ equals to one, and we can obtain the following equation:
		\begin{eqnarray}
			\label{確率密度の変数変換(2)}
			p(\mathbf{x} | \mbox{\boldmath $\theta$}^{(c)},G_m) = p_\mathbf{e}^{(c)}(\mathbf{e}^{(c)})
		\end{eqnarray}
		According to the assumption in LiNGAM that the $e_i$ are mutually independent,
		\begin{eqnarray}
			\label{eの確率密度の分解}
			p_\mathbf{e}^{(c)}(\mathbf{e}) &=& \prod_{i = 1}^{n} p_i^{(c)} ( e_i^{(c)} ).
		\end{eqnarray}
		Then, the probability density of $\mathbf{x}$ within each class $c$, using equations (\ref{LiNGAM混合モデルの式}), (\ref{確率密度の変数変換(2)}), and (\ref{eの確率密度の分解}), is given by
		\begin{eqnarray}
			\label{xの確率密度（各クラス）}
		      && \nonumber p(\mathbf{x} | \mbox{\boldmath $\theta$}^{(c)},G_m) \\
			&& \nonumber = \prod_{i = 1}^{n} p_i^{(c)} ( e_i^{(c)} ) \\
			&& \nonumber = \prod_{i = 1}^{n} p_i^{(c)} ( x_i - \mu_i^{(c)} - \sum_{k^{(c)}(j) < k^{(c)}(i)}^{} b_{ij}^{(c)}
			 (x_j - \mu_j^{(c)})). \\
		\end{eqnarray}
		We can now specify $p(\mathbf{D}|\mathbf{\Theta}, G_m)$ with the assumption of IID data and equations (\ref{確率密度の混合}) and (\ref{xの確率密度（各クラス）}):
		\begin{eqnarray}
			\label{Dの確率密度}
			&& p(\mathbf{D}|\mathbf{\Theta},G_m) = \prod_{s=1}^{N} p(\mathbf{x}^s | \mathbf{\Theta}, G_m) \\
			\label{xの確率密度}
			\nonumber && p(\mathbf{x}|\mathbf{\Theta},G_m) \\
			\nonumber && =\sum_{c = 1}^{l} \biggl\{ p(\mathbf{x} | \mbox{\boldmath $\theta$}^{(c)},G_m) \biggr\} P(c) \\ 
			\nonumber && =\sum_{c = 1}^{l} \biggl\{ \prod_{i = 1}^{n} p_i^{(c)} \bigl( x_i - \mu_i^{(c)} - \sum_{k^{(c)}(j) < k^{(c)}(i)}^{} b_{ij}^{(c)} (x_j - \mu_j^{(c)})\bigr)\biggr\} \\
					   &&  \times P(c)
		\end{eqnarray}

	\subsection{Priors for parameters and hyperparameters}
	\label{事前分布}
		We use multinomial distribution for $P(c)$ in Equation (\ref{確率密度の混合}). The distribution represents, for $N$ independent trials, each of which leads to one of the possible events $c = 1, \dots, l$, the number of times that each event occurs ($\mathbf{z} = [z^{(1)}, \dots, z^{(l)}]$). $w^{(c)}$ is the probability of the occurrence of each event. Generally, the probability mass function of a multinomial distribution is as follows:
		\begin{eqnarray}
			\label{多項分布（一般）}
			P(\mathbf{z}) = \frac{N!}{\prod_{c=1}^{l} z^{(c)}!} \prod_{c=1}^{l} (w^{(c)})^{z^{(c)}}
		\end{eqnarray}
		where $w^{(c)} > 0$ and $\sum_{c = 1}^{l} w^{(c)} = 1$.
		When we use the above distribution in Equation (\ref{確率密度の混合}), the number of trials $N$ is one. Then, $\mathbf{z}$ can be regarded as the indicator vector and corresponds to variable $c$. For example, if $\mathbf{x}$ belongs to class $2$ (i.e., $c = 2$), $z^{(2)} = 1$ and the other components are $0$.
		From the above, we can use a multinomial distribution for $P(c)$ in the following form:
		\begin{eqnarray}
			\label{多項分布（使用）}
			P(c) = w^{(c)}
		\end{eqnarray}

		We use a Dirichlet distribution for the parameters of the multinomial distribution ($w^{(1)}, \dots, w^{(l)}$) because the former is a conjugate prior for the latter and is typically used in similar contexts.
		The density function of the Dirichlet distribution is as follows:
		\begin{eqnarray}
			\label{ディリクレ分布}
			p(\mathbf{w}) = \frac{\Gamma(\sum_{c=1}^{l} a^{(c)})}{\prod_{c=1}^{l} \Gamma(a^{(c)})} \prod_{c=1}^{l} (w^{(c)})^{a^{(c)}-1}
		\end{eqnarray}
		where $\mathbf{w} = w^{(1)}, \dots, w^{(l)}$ and $a^{(c)}(>0)$ are concentration parameters.
		We can generate a random vector $\mathbf{w}$ by normalizing independent gamma random variables $\gamma^{(1)}, \dots , \gamma^{(l)}$ with shape parameters $a^{(c)}$ (concentration parameters in the Dirichlet distribution) and scale parameter 1  \cite{Devroye86book}:
		\begin{eqnarray}
			w^{(c)} &=& \frac{\gamma^{(c)}}{\sum_{c=1}^{l}\gamma^{(c)}} \\
			p(\gamma^{(c)}) &=& \frac{1}{\Gamma(a^{(c)})} (\gamma^{(c)})^{a^{(c)}-1}{\rm exp}(\gamma^{(c)})
		\end{eqnarray}

		We use Gaussian distributions for $b_{ij}^{(c)}$ and $\mu_i^{(c)}$ with zero and $\varphi^{(c)}$ as their means and $v^2$ and $\tau^2$ as their variances respectively.
		\begin{eqnarray}
			p(b_{ij}^{(c)}) &=& \frac{1}{\sqrt{2 \pi v^2}}{\rm exp}\biggl\{ -\frac{(b_{ij}^{(c)})^2}{2v^2}\biggr\} \\
			p(\mu_i^{(c)}) &=& \frac{1}{\sqrt{2 \pi \tau^2}}{\rm exp}\biggl\{ -\frac{(\mu_i^{(c)} - \varphi^{(c)})^2}{2\tau^2}\biggr\}
		\end{eqnarray}
		We heuristically determine the value of $\varphi$ .
		We use the result of the Gaussian mixture model estimation and an expectation-maximization (EM) algorithm \cite{Bishop2006PRML} for $\varphi$ because the greater the number of latent classes, the more difficult it is to estimate $\mu_i^{(c)}$.
		The EM algorithm is a method to compute maximum likelihood solutions for models with latent variables. In the algorithm, parameters and responsibilities (conditional probabilities of latent variables given the data) are updated in turn until the change in the likelihood function falls below some threshold.

		We use inverse gamma distributions \cite{cook2008inverse} with shape parameters $\alpha$, $\eta$, and $\chi$ and scale parameters $\beta$, $\zeta$, and $\epsilon$ for $(\sigma_i^{(c)})^2$, $\lambda_i^{(c)}$, and $v^2$, respectively.
		\begin{eqnarray}
			\label{逆ガンマ分布}
			p((\sigma_i^{(c)})^2) &=& \frac{\beta^\alpha}{\Gamma(\alpha)}(\sigma_i^{(c)})^{2(-\alpha-1)}{\rm exp}\biggl\{\frac{-\beta}{(\sigma_i^{(c)})^2}\biggr\} \\
			p(\lambda_i^{(c)}) &=& \frac{\zeta^\eta}{\Gamma(\eta)}(\lambda_i^{(c)})^{-\eta-1}{\rm exp}\biggl(\frac{-\zeta}{\lambda_i^{(c)}}\biggr) \\
			p(v^2) &=& \frac{\epsilon^\chi}{\Gamma(\chi)}v^{2(-\chi-1)}{\rm exp}\biggl(\frac{-\epsilon}{v^2}\biggr)
		\end{eqnarray}
		We can generate inverse gamma random variables using gamma random variables. If a variable $X$ has a gamma distribution with shape parameter $\alpha$ and scale parameter $\beta$, $Y = 1/X$ has an inverse gamma distribution with shape parameter $\alpha$ and scale parameter $1/\beta$ \cite{cook2008inverse}.

		We determine the value of $a^{(c)}$, $\alpha$, $\beta$, $\eta$, $\zeta$, $\chi$, $\epsilon$, and $\tau$ arbitrarily.
		From the above, we can compute $P(\mathbf{D}|G_m)$ in Equation (\ref{周辺尤度}) using ordinary Monte Carlo sampling \cite{Bishop2006PRML} to compute the integral.

\section{Simulation}
\label{評価実験}
		In this simulation, we assume for simplicity that we know of the existence of a causal connection between two observed variables, and that its direction is the same for all classes. We can thus estimate which direction is true ($x_1^{(c)} \rightarrow x_2^{(c)}$ or $x_1^{(c)} \leftarrow x_2^{(c)}$).
		We generated 1,000 datasets under every combination of sample size ($N = 50, 100, 500$) and the number of classes ($l = 2, 4, 6$).

		The data within each class were generated by the basic LiNGAM (Equation(\ref{LiNGAMモデルの式})), following which we mixed them.
		In all datasets, the true model was $x_1^{(c)} \rightarrow x_2^{(c)}$ and the connection strength was $b_{ij}^{(c)} = [-1.5, -0.5] \cap [0.5, 1.5]$.
		The distributions of external influences $e_i^{(c)}$ were randomly selected from the following three non-Gaussian distributions with one variance: Laplace, Uniform ([$-\sqrt{3},\sqrt{3}$]) and Student-t (with five degrees of freedom).

		With regard to the priors we did not use any prior information about $G_m$, and thus $P(G_m) = \frac{1}{2}$. We randomly selected the value of $a^{(c)}$ from among $3$, $5$, and $7$. Our choice of values for the rest of the parameters was $\alpha = 3$, $\beta = 3$, $\eta = 3$, $\zeta= 3$, $\chi= 3$, $\epsilon = 3$, and $\tau = 0.5$.

		We selected the number of classes $l$ as follows. We calculated the log-marginal likelihoods of the two models for all numbers of classes and selected the class with the largest log-marginal likelihood. The maximum number of classes tested was $2{\rm log}N$. This was motivated by the fact that when the sample size $N$ approaches infinity in a Dirichlet process mixture model, the number of classes $l$ approaches ${\rm log}N$ \cite{antoniak1974DP}.

		We compared our method with the existing method \cite{Shimizu07ICONIP} to determine the one that can correctly estimate causal directions more times given the above datasets.
 		However, we use \cite{Palmer08ICASSP} in the existing method \cite{Shimizu07ICONIP} instead of \cite{Mollah-mixtures}.
		We show the simulation results in Table \ref{result}. The numbers in the tables denote the number of times each method correctly estimated causal directions. As shown in the table, our method has more correct answers than the existing method.
		\begin{table}
		\begin{center}
		\caption{Simulation results}
			\setlength{\doublerulesep}{1.0pt}
			\small
			{\setlength{\tabcolsep}{5pt}
			\begin{tabular*}{\columnwidth}{@{\extracolsep{\fill}}ll|lllllll}
												&					& \multicolumn{3}{c}{Sample size} \\
												&			& 50		& 100	& 500	\\ \hline \hline
								Our method		&	$l=2$	& 913	& 947	& 981	\\
												&	$l=4$	& 908	& 937	& 973	\\
												&    	$l=6$	& 922	& 957	& 967	\\ \hline
								Existing method	&	$l=2$	& 649	& 657	& 684	\\
												&    	$l=4$	& 663	& 655	& 729	\\
												&    	$l=6$	& 646	& 700	& 762	\\ \hline
			\end{tabular*}
		}
		\label{result}
		\vspace{-6mm}\\
		\end{center}
		\end{table}
%

\section{Conclusion}
\label{結論}
		In this paper we proposed a new estimation approach for discovering causal structure in the presence of latent classes.
		In a simulation run using artificial data, our method correctly estimated more causal directions than the existing method.
		Our plan for future research is to evaluate our method on a wide variety of real datasets.

\bibliography{jsai2014_ref}

\end{document}